\documentclass[conference]{IEEEtran}
\IEEEoverridecommandlockouts

\usepackage{hyperref} 

\usepackage{cite}
\usepackage{amsmath,amssymb,amsfonts}
\usepackage{algorithmic}
\usepackage{graphicx}
\usepackage{textcomp}

\usepackage{url}
\usepackage{svg}
\usepackage{numprint}
\npdecimalsign{.}

\usepackage{xstring}
\usepackage{multirow}
\usepackage{caption}
\usepackage{subcaption}

\usepackage{xcolor,colortbl}

\def\BibTeX{{\rm B\kern-.05em{\sc i\kern-.025em b}\kern-.08em
    T\kern-.1667em\lower.7ex\hbox{E}\kern-.125emX}}
\begin{document}

\title{A Comprehensive Analysis of Tokenization and Self-Supervised Learning in End-to-End Automatic Speech Recognition applied on French Language
}

\author{\IEEEauthorblockN{Thibault Ba\~{n}eras-Roux}
\IEEEauthorblockA{\textit{Nantes University} \\
\textit{LS2N}\\
Nantes, France}
\and
\IEEEauthorblockN{Mickael Rouvier}
\IEEEauthorblockA{\textit{Avignon University} \\
\textit{LIA}\\
Avignon, France}
\and
\IEEEauthorblockN{Jane Wottawa}
\IEEEauthorblockA{\textit{Le Mans University} \\
\textit{LIUM}\\
Le Mans, France}
\and
\IEEEauthorblockN{Richard Dufour}
\IEEEauthorblockA{\textit{Nantes University} \\
\textit{LS2N}\\
Nantes, France}
}

\maketitle

\begin{abstract}
The performance of end-to-end automatic speech recognition (ASR) systems enables their increasing integration into numerous applications. While there are various benefits to such speech-to-text systems, the choice of hyperparameters and models plays a crucial role in their performance. Typically, these choices are determined by considering only the character (CER) and/or word error rate (WER) metrics. However, it has been shown in several studies that these metrics are largely incomplete and fail to adequately describe the downstream application of automatic transcripts. In this paper, we conduct a qualitative study on the French language that investigates the impact of subword tokenization algorithms and self-supervised learning models from different linguistic and acoustic perspectives, using a comprehensive set of evaluation metrics. 
\end{abstract}


\section{Introduction}

Automatic Speech Recognition (ASR) technology is integral to various applications, including transcription services, voice assistants, and automated captioning. Its ability to convert spoken language into written text has significantly enhanced the accessibility and usability of audio content. With the ever-increasing demand for precise and efficient ASR systems, researchers are continuously exploring innovative methods to enhance their performance.

ASR models heavily rely on tokenization as a foundational element in the transcription process. Traditionally, word tokenization segments text into individual words using predefined delimiters like spaces and punctuation marks. ASR systems predict these tokens with the assistance of a decoder. Modern ASR systems employ a more sophisticated tokenization approach, segmenting words into smaller units known as subwords. This finer tokenization enhances the system's ability to handle out-of-vocabulary (OOV) words and reduces vocabulary size.  Among the prominent tokenization approaches in use, we can mention Byte-Pair Encoding (BPE)~\cite{sennrich2016neural}
or SentencePiece~\cite{kudo2018sentencepiece}.

Another critical aspect shaping the advancement of ASR systems is Self-Supervised Learning (SSL). Notably, the development of wav2vec~\cite{baevski2020wav2vec} and~\cite{hsu2021hubert}  has significantly bolstered the acoustic generalization capabilities of ASR systems. These SSL models are trained without the need for manual annotations, leveraging vast amounts of audio data to generate speech representations known as {\it embeddings}. These embeddings serve as concise representations of speech segments, capturing essential acoustic characteristics from the speech data. When integrated into ASR systems, they substantially enhance adaptability to various speaking styles, accents, and background noise, resulting in more robust and accurate speech recognition.

However, the impact of these parameters remains relatively unexplored within ASR research, particularly on end-to-end ASR systems, eliminating the need for intermediate representations or separate processing stages. While these architectures are gaining importance, our understanding of them is still in its early stages. Evaluation often focuses predominantly on metrics such as Word Error Rate (WER), while broader implications of tokenization and SSL choices on transcription quality are seldom examined or rigorously investigated. This paper aims to fill this gap. Building upon previous work~\cite{singh2021comparative}, we propose a comprehensive study examining the effects of tokenizers and SSL models on lexical, acoustic, and semantic metrics~\cite{zhang2019bertscore, kim2021semantic, roux2022qualitative} specifically tailored to the French language.

In this paper, we make the following contributions:
\begin{itemize}
    \item We establish that a reduced vocabulary enhances the generalization capabilities of ASR systems.
    \item We provide compelling evidence for the effectiveness of using the Unigram tokenizer with a reduced vocabulary, particularly in the context of French language ASR, resulting in improved performance.
    \item We demonstrate that evaluation criteria for metrics have a discernible impact at the system level. Consequently, a system deemed optimal by WER may not necessarily be the best from a human, semantic, or other perspectives.
    \item Our results suggest that the use of phonetically adapted tokens does not yield performance improvements compared to traditional tokenization methods.
\end{itemize}

The paper is organized as follows. In Section~\ref{sec:methodology}, we describe the methodology employed for evaluating the different tokenizers and SSL models, and discuss the evaluation metrics. Then, we study how the choice of self-supervised models influences ASR performance in Section~\ref{sec:ssl-section} and we carry out an analysis of the impact of tokenizer hyperparameters in Section~\ref{sec:tokenization}. In Section~\ref{sec:metric-discrepancy}, we discuss the difficulties in clearly assessing the ASR performance, with the nature of the evaluation metric influencing the reported quality of a transcription system. We finally conclude in~Section~\ref{sec:conclusion}.

\begin{figure}[h]
\centering
\includegraphics[width=0.48\textwidth]{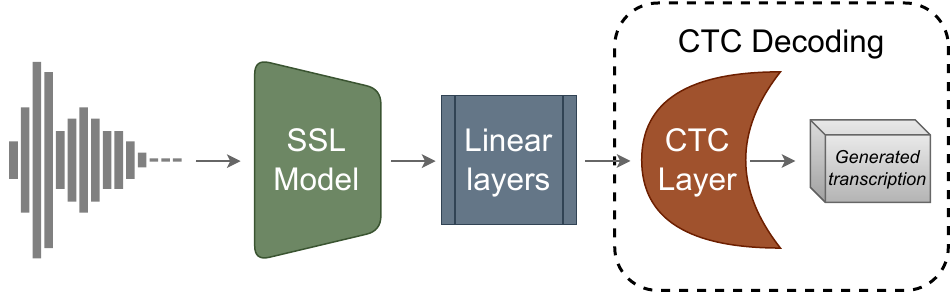}
\caption{Architecture of the Automatic Speech Recognition systems used in this study.}
\label{fig:archicture}
\end{figure}

\section{Study Methodology}
\label{sec:methodology}


In this section, we detail our study setup. Firstly, we discuss two key components, including their studied approaches: tokenization strategies (Section~\ref{sec:tok}) and self-supervised learning models (Section~\ref{sec:ssl}). Then, we include a description of corpora (Section~\ref{sec:corpora}) and evaluation metrics (Section~\ref{sec:metrics}) to study various performance aspects. Finally, we describe the used end-to-end ASR system (Section~\ref{sec:asr}).

\subsection{Tokenization strategies}
\label{sec:tok}

\newcolumntype{C}[1]{>{\centering\arraybackslash}m{#1}}

\begin{table*}[]
\centering
\begin{tabular}{|C{2.3cm}|C{6cm}|C{1.75cm}|C{1.75cm}|C{1.75cm}|}
\hline
\textbf{Tokenizer} & \textbf{Description} & \textbf{Word split} & \textbf{Operation} & \textbf{Scoring} \\ \hline
BPE & Initializes the vocabulary with individual characters and iteratively merges the most frequent token pairs until the desired vocabulary size is achieved. & Yes & Merge & Frequency \\ \hline
WordPiece & Similar to BPE, it selects the token pair that maximizes the likelihood of the training data, rather than choosing the most frequent pair. & Yes & Merge & Likelihood \\ \hline
Unigram & Initializes the vocabulary with a large number of tokens and then systematically reduces the size of the vocabulary by iteratively trimming each token. & No & Trim & Likelihood \\ \hline\hline
SentencePiece & Tool that employs BPE or Unigram algorithms without segmenting the training data into words. & No & Both & Both \\ \hline
\end{tabular}
\caption{Overview of the most commonly used subword tokenizers.}
\label{tab:tokenizer-overview}
\end{table*}

Tokenization techniques have undergone significant evolution, employing diverse algorithms to process language. Tokenization involves breaking raw text into smaller units known as tokens, which can represent words, subwords, or characters. In end-to-end ASR, the prevailing approach computes a sequence of token probabilities for individual speech segments, generating transcriptions using the Connectionist Temporal Classification (CTC) framework.

Among the most used tokenization strategies, Byte-Pair Encoding (BPE) tokenization~\cite{sennrich2016neural} segments text into subwords using a vocabulary initialized with characters and expanded through an iterative merging of the most frequent token pairs. WordPiece~\cite{schuster2012japanese} shares similarities with BPE but adopts a likelihood-based merging approach. Unigram~\cite{kudo2018subword}, on the other hand, focuses on trimming a large vocabulary using loss-based criteria. Additionally, SentencePiece~\cite{kudo2018subword} implements both Unigram and BPE but does not pre-tokenize sentences into words. These subword tokenization strategies are summarized in Table~\ref{tab:tokenizer-overview}.


We employ various tokenization methods, including character, BPE, SentencePiece, and Unigram, while varying the vocabulary size for subword tokenization. 
Given that speech is the primary modality in ASR, investigating the use of grapheme-based linguistic tokenization may be pertinent. Unlike tokens chosen based on the co-occurrence frequency of characters and subwords, grapheme-based tokenization fully respects linguistic and acoustic characteristics.  To this end, we train systems with a vocabulary based on 144 graphemes\footnote{aa, ae, aen, ai, aï, ail, aim, ain, am, an, aon, aou, au, aw, ay, aye, bb, ca, cc, cca, cce, cch, cci, cco, ccu, ccueil, ccy, ce, ch, ci, co, cqu, ct, cu, cueil, cy, dd, ds, ea, ean, eau, ect, ed, ee, ée, ef, ei, eil, eim, ein, em, emmm, en, enn, ent, er, es, eu, eû, ew, ez, ff, ga, ge, geu, geü, gg, gge, ggi, gh, gi, gn, go, gt, gu, gua, gue, guë, güe, gui, ign, iil, il, ill, illaire, ille, illier, im, imm, imma, imme, immi, immo, immu, in, ing, ll, lle, mm, mn, nn, oa, oe, oi, oil, om, on, ou, ph, pp, ps, pt, qu, qua, qui, rh, rr, rrh, sc, sca, sce, sch, sci, sco, scu, scy, ss, th, tia, tie, tiel, tien, tient, tieuse, tieux, tion, tt, tz, uil, um, un, uy, ym, yn} (character string transcribing a phoneme), compiled by cross-referencing various teaching resources. With BPE, the vocabulary can be initialized with a specific set of tokens, not just characters. Thus, we initialize BPE tokenizers with graphemes.



\subsection{Self-supervised learning (SSL) models}
\label{sec:ssl}

We employed multiple SSL models, each trained on diverse datasets and languages. Specifically, we used LeBenchmark large models~\cite{evain2021task}, which are wav2vec 2.0 models trained on different amounts of French data: 1,000 hours (w2v2-FR-1k), 3,000 hours (w2v2-FR-3k), and 7,000 hours (w2v2-FR-7k). This allowed us to assess ASR performance across various amounts of training data. Additionally, we included the classic wav2vec 2.0 model (w2v2-EN-53k)~\cite{baevski2020wav2vec}, trained on 53,000 hours of English data, to investigate its transfer-learning capabilities to another language. Furthermore, we incorporated an XLSR model (w2v2-xlsr)~\cite{babu2021xls}, which is a wav2vec 2.0 model trained on a diverse dataset comprising 53 languages, including French. This enabled us to examine the effects of cross-lingual training on ASR performance. Using these SSL models, our goal was to gain a comprehensive understanding of the impact of training data size and monolingual/multilingual training, focusing on identifying the most effective approaches for developing end-to-end ASR systems.

\subsection{Corpora}
\label{sec:corpora}

The relationship between spoken and written French is intriguing due to the relatively large presence of silent letters, which can induce distinctive behavior of semantic and lexical metrics at the word and character level. Therefore, all end-to-end ASR systems have been trained to process French using ESTER
1~\cite{galliano2006corpus} and ESTER 2~\cite{galliano2009ester}, EPAC~\cite{esteve2010epac}, ETAPE~\cite{gravier2012etape} and REPERE~\cite{giraudel2012repere} train corpora. Collectively, these corpora represent approximately 356 hours of audio, comprised of radio and television broadcast data.

Our comprehensive analysis is based on the French REPERE test corpus, which corresponds to 10 hours of speech.

\subsection{Automatic Speech Recognition systems}
\label{sec:asr}

In this study, we set up 28 end-to-end ASR systems based on the Speechbrain toolkit~\cite{ravanelli2021speechbrain}.
All the ASR systems incorporate an SSL model, a Deep Neural Network (DNN) layer composed of three linear layers and a CTC layer, as shown in Figure~\ref{fig:archicture}. These systems are trained for 10 epochs using CTC loss on the corpora described in Section~\ref{sec:corpora} with a lower learning rate for the SSL model. For inference, the transcription is generated with \textit{best path decoding}. For reproducibility, settings are detailed in our GitHub code repository\footnote{\href{https://github.com/thibault-roux/systems-analysis}{{https://github.com/thibault-roux/systems-analysis}}}.

\begin{table}[h!]
\centering
\begin{tabular}{|c|c|c|c|c|c|}
\hline
\textbf{SSL model} & \textbf{WER} & \textbf{CER} & \textbf{SemDist} & \textbf{UWER} & \textbf{PhonER} \\ \hline\hline 
\textbf{w2v2-FR-1K} & 18.94 & 7.63 & 12.52 & 77.42 & 6.26 \\ \hline 
\textbf{w2v2-FR-3K} & 17.16 & 6.87 & 11.20 & 76.84 & 5.44 \\ \hline 
\textbf{w2v2-FR-7k} & \textbf{16.56} & \textbf{6.72} & \textbf{10.45} & \textbf{75.19} & \textbf{5.29} \\ \hline\hline 
\textbf{w2v2-xlsr} & 21.48 & 8.59 & 14.47 & 78.66 & 7.03 \\ \hline\hline 
\textbf{w2v2-EN-53k} & 36.41 & 13.67 & 23.62 & 89.83 & 12.63 \\ \hline 
\end{tabular}
\caption{Performance of ASR systems using a character tokenizer and different SSL models (French models with w2v2-FR-1k, w2v2-FR-3k, and w2v2-FR-7k; English model with w2v2-EN-53k; multilingual model with w2v2-xlsr).}
\label{tab:scores-ssl}
\end{table}

\subsection{Evaluation metrics}
\label{sec:metrics}

Instead of focusing solely on the classical WER metric, we examine various aspects of automatic transcriptions using metrics that evaluate lexical, semantic, and acoustic levels.

For the lexical aspect, we consider classical metrics such as \textbf{Word Error Rate (WER)} and \textbf{Character Error Rate (CER)}. Additionally, inspired by the Individual Word Error Rate~\cite{mdhaffar2019qualitative} and aiming to study the generalization ability of ASR systems, we developed the \textbf{Unseen Word Error Rate (UWER)}. The UWER measures the accuracy of transcribed words specifically for those absent from the training corpora but present in the test set, providing a valuable assessment of the system's ability to generalize to unseen vocabulary. 

At the semantic level, we employ the \textbf{SemDist}~\cite{kim2021semantic} metric, which computes the cosine similarity between embeddings of the reference and hypothesis obtained at the sentence level. In our experiments, we utilized a sentence embedding model\footnote{\href{https://huggingface.co/dangvantuan/sentence-camembert-large}{https://huggingface.co/dangvantuan/sentence-camembert-large}} (SentenceBERT~\cite{reimers2019sentence}) based on CamemBERT~\cite{martin2020camembert}, a French pre-trained BERT version. This metric had the strongest correlation with human perception in a previous study~\cite{baneras2023hats}.

In addition to text transcripts derived from speech, we also consider an acoustic metric: the {\bf Phoneme Error Rate (PhonER)}, which involves computing the Levenshtein distance between reference and hypothesis sequences of phonemes both obtained using an automatic grapheme-to-phoneme converter\footnote{\href{https://github.com/Remiphilius/PoemesProfonds}{https://github.com/Remiphilius/PoemesProfonds}}.

\section{Impact of SSL models}
\label{sec:ssl-section}

In this section, we reproduce previous results~\cite{evain2021benchmark} on how the language (Section~\ref{sec:ssl-training-language}) and the size (Section~\ref{sec:ssl-training-size}) of the training data used by SSL models affect the end-to-end ASR system's performance and deepen this analysis by using several metrics. To ensure a fair comparison between SSL models, all the results in this section use a character tokenizer. Table~\ref{tab:scores-ssl} presents the performance obtained by our end-to-end ASR system using various SSL model configurations. 

\subsection{Impact of training language}
\label{sec:ssl-training-language}

As depicted in Table~\ref{tab:scores-ssl}, SSL models pre-trained on French data (\textit{w2v2-FR-*}) consistently demonstrate superior performance across metrics. Conversely, the English-based system (\textit{w2v2-EN-53k}), despite having the largest training dataset, exhibits a relatively high Word Error Rate (WER) of 36.41\%. In contrast, the ASR system trained on the target language achieves a substantially improved WER of 16.52\% using the same character tokenizer (\textit{w2v2-FR-7k}). This highlights the significance of training SSL models on the target language to acquire language-specific knowledge crucial for accurate transcription.

When fine-tuning an SSL model trained on a diverse dataset that includes the target language (\textit{w2v2-xslr}), we observe a performance drop compared to the monolingual French system, resulting in a WER of 21.48\%. However, this multilingual system still outperforms the English-based ASR system. It is important to note that in multilingual training, there is a risk that language-specific information may become overwritten, diluted, or averaged by the inclusion of other languages.

\begin{table*}[ht!]
\centering
\begin{tabular}{|c|c|c|c|c|c|c||c|}
\hline
\textbf{Tokenizer} & \textbf{\# Token} & \textbf{WER} & \textbf{CER} & \textbf{SemDist} & \textbf{UWER} & \textbf{PhonER} & \textbf{Avg. token} \\ \hline\hline 
 \multirow{5}{*}{\textbf{BPE}} & 1000  & 15.98 & 7.00 & 10.08 & 78.74 & 5.72 & 1.92 \\ \cline{2-8} 
 & 750 &  15.33 & 6.67 & 9.41 & 76.67 & 5.31 & 2.05 \\ \cline{2-8} 
 & 500 &  15.57 & 6.73 & 9.61 & 76.43 & 5.38 & 2.28 \\ \cline{2-8} 
 & 250 & 15.16 & 6.45 & 9.43 & 74.11 & 5.05 & 2.75 \\ \cline{2-8} 
 & 150  & 15.47 & 6.46 & 9.47 & 74.77 & 5.10 & 3.20 \\ \hline 
\multirow{4}{*}{\textbf{BPE with graphemes}}  & 1000  & 15.74 & 6.62 & 9.97 & 77.25 & 5.40 & 2.76 \\ \cline{2-8}
 & 750  & 15.98 & 6.63 & 10.03 & 77.58 & 5.47 & 2.81 \\ \cline{2-8}
 & 500  & 15.64 & 6.59 & 9.77 & 76.51 & 5.34 & 2.93 \\ \cline{2-8}
 & 250  & 15.74 & 6.55 & 9.73 & 75.77 & 5.18 & 3.10 \\ \hline
\multirow{5}{*}{\textbf{SentencePiece}} & 1000  & 15.78 & 6.87 & 9.76 & 77.83 & 5.14 & 1.88 \\ \cline{2-8} 
& 750  & 15.59 & 6.76 & 9.39 & 76.18 & 5.35 & 2.03 \\ \cline{2-8}
 & 500  & 15.51 & 6.66 & 9.55 & 76.43 & 5.33 & 2.26 \\ \cline{2-8} 
 & 250  & 15.70 & 6.74 & 9.75 & 74.52 & 5.37 & 2.75 \\ \cline{2-8} 
 & 150  & 15.56 & 6.57 & 9.52 & 74.52 & 5.58 & 3.29\\ \hline
 \multirow{5}{*}{\textbf{Unigram}} & 1000  & 15.49 & 6.68 & 9.57 & 78.91 & 5.37 & 1.88 \\ \cline{2-8} 
& 750  & 15.29 & 6.55 & 9.34 & 76.67 & 5.23 & 2.03 \\ \cline{2-8}
 & 500  & 15.54 & 6.70 & 9.57 & 76.26 & 5.29 & 2.26 \\ \cline{2-8}
 & 250  & 15.58 & 6.65 & 9.44 & 73.53 & 5.23 & 2.77  \\ \cline{2-8}
 & 150  & \textbf{15.07} & \textbf{6.36} & \textbf{9.33} & \textbf{73.12} & \textbf{4.90} & 3.33 \\ \hline
 \textbf{Character} & -  & 16.56 & 6.72 & 10.45 & 75.19 & 5.29 & 4.88 \\ \hline 
\end{tabular}
\caption{Performance of ASR systems using different tokenizers (BPE, character, graphemes, SentencePiece and Unigram).}
\label{tab:scores-tokenizer}
\end{table*}

\subsection{Impact of training data size}
\label{sec:ssl-training-size}

We now narrow our focus to the analysis of the French models only (\textit{w2v2-FR-*}), as presented in Table~\ref{tab:scores-ssl}, to examine the impact of SSL training data size. Our analysis reveals a clear and direct correlation between the size of the training data and improved performance across all considered metrics. Increasing the training data size enables the model to learn more comprehensive representations and better capture a wider range of acoustic and linguistic variations.

\section{Impact of Tokenization Strategies}
\label{sec:tokenization}

In this section, we explore how tokenization algorithms can impact the performance assessment of ASR systems. We begin by comparing subword units and character tokenization (Section~\ref{s:sub}). 
Then, we assess the use of graphemes as subword units to determine the most suitable subword unit for optimizing ASR system performance (Section~\ref{s:grap}).

Table~\ref{tab:scores-tokenizer} presents the performance of end-to-end ASR systems trained with different tokenization strategies. The last column of the table, {\it Avg. token}, represents the average sub-word units per word for each tokenizer on the test dataset. To ensure a fair comparison between tokenization strategies for all ASR systems, we employed the SSL model {\it w2v2-FR-7k}, known for its superior performance. Notably, the system utilizing the Unigram tokenizer with a fixed vocabulary of 150 consistently achieved the best results across various metrics, including WER, CER, SemDist, UWER, and PhonER.

\subsection{Subword units vs Character tokenization}
\label{s:sub}

We observe in Table~\ref{tab:scores-tokenizer} that subword unit tokenizers (BPE, SentencePiece, and Unigram) consistently outperform character tokenization since this tokenization neglects linguistic and acoustic intricacies in speech. In contrast, subword unit tokenizers capture more nuanced and contextually relevant information, which explains their better performance across all metrics.

\subsection{Influence of graphemes}
\label{s:grap}

It is noteworthy that the \textit{BPE with grapheme} tokenizer consistently yields inferior results compared to other subword unit tokenizers. Despite its intention to integrate knowledge about acoustics and linguistics, end-to-end ASR models struggle to effectively utilize this information, resulting in suboptimal outcomes.

Table~\ref{tab:perc-grapheme} displays the percentage of graphemes included in the vocabulary of other tokenizers. An interesting finding is that the best-performing system has the lowest percentage of graphemes. This observation, coupled with the lower relative performance of systems using graphemes, suggests that subword units align more closely with linguistic elements than with acoustics.

\begin{table}[]
\centering
\begin{tabular}{|c|c|c|c|}
\hline
\textbf{\# Token} & \textbf{SentencePiece} & \textbf{BPE} & \textbf{Unigram} \\ \hline \hline
250 & 17.24\% & 17.24\% & 11.03\% \\ \hline
500 & 21.38\% & 21.38\% & 17.93\% \\ \hline
750 & 24.14\% & 24.14\% & 22.07\% \\ \hline
1000 & 27.59\% & 28.97\% & 23.45\% \\ \hline
\end{tabular}%
\caption{Percentages of graphemes included in the vocabulary of different tokenizers.}
\label{tab:perc-grapheme}
\end{table}

\section{Metrics discrepancy}
\label{sec:metric-discrepancy}


Despite the 150-size Unigram tokenizer outperforming others for all metrics (as demonstrated in Table~\ref{tab:scores-tokenizer}), metrics fail to establish a consistent ranking between systems. For instance, for SentencePiece, the best system has a vocabulary size of 500 according to WER, 150 according to CER and UWER, and PhonER, and 750 according to SemDist. Table~\ref{t:corrs} illustrates the Spearman correlation between metrics at the system level, revealing that the indicated hierarchy can vary significantly.

The discrepancies between metrics pose challenges in determining a clear best system because different metrics offer conflicting rankings. This inconsistency prompts questions about the relevance of standard metrics like WER for accurately evaluating system performance. Previous research~\cite{baneras2023hats} has already shown that metrics do not equally correlate with human perception. In the context of French and across a range of metrics assessing aspects like lexical accuracy, semantics, and phonetics, it was observed that, at the utterance level, WER had one of the lowest correlations with human perception, while SemDist, using sentence embeddings, exhibited the strongest correlation. In our study, these differences underscore that ASR metrics can yield varying assessments of performance at the system level, which is a first, to our knowledge. 


\begin{table}[!htb]
\centering
\begin{tabular}{c|c|c|c|c|c|}
\cline{2-6}
\textbf{} & \textbf{WER} & \textbf{CER} & \textbf{SemDist} & \textbf{UWER} & \textbf{PhonER} \\ \hline
\multicolumn{1}{|c|}{\textbf{WER}} & \cellcolor[gray]{0.9} & \cellcolor[gray]{0.9} & \cellcolor[gray]{0.9} & \cellcolor[gray]{0.9} & \cellcolor[gray]{0.9} \\ \hline
\multicolumn{1}{|c|}{\textbf{CER}} & 0.55 & \cellcolor[gray]{0.9} & \cellcolor[gray]{0.9} & \cellcolor[gray]{0.9} & \cellcolor[gray]{0.9} \\ \hline
\multicolumn{1}{|c|}{\textbf{SemDist}} & 0.87 & 0.45 & \cellcolor[gray]{0.9} & \cellcolor[gray]{0.9} & \cellcolor[gray]{0.9} \\ \hline
\multicolumn{1}{|c|}{\textbf{UWER}} & 0.34 & 0.45 & 0.47 & \cellcolor[gray]{0.9}{\cellcolor[gray]{0.9} } & \cellcolor[gray]{0.9} \\ \hline
\multicolumn{1}{|c|}{\textbf{PhonER}} & 0.63 & 0.76 & 0.61 & 0.80 & \cellcolor[gray]{0.9} \\ \hline
\end{tabular}
\caption{Spearman correlation of metrics at system level.} \label{t:corrs}
\end{table}

\section{Conclusion and perspectives}
\label{sec:conclusion}

In this paper, we conducted a thorough analysis of two pivotal factors influencing ASR system performance: tokenization strategy and self-supervised learning (SSL) models. Our findings shed light on the intricate relationship between these components and various language aspects, offering valuable insights for the speech community.

Regarding tokenization, our analysis unveiled that systems with larger vocabulary sizes encountered challenges in generalizing to out-of-vocabulary (OOV) words. Conversely, character tokenization excelled in terms of Character Error Rate (CER) but faced difficulties in maintaining lexical accuracy and word boundaries.

In the realm of SSL models, we corroborate the conclusions of previous works~\cite{evain2021benchmark} by observing a direct correlation between training data size and improved ASR system performance across all metrics. Larger SSL model training datasets in the target language facilitate better generalization and enhanced representation learning, resulting in overall improved performance. Additionally, our study underscores the significance of pre-training SSL models on the target language, as models not specifically trained on it exhibited performance limitations due to the lack of language-specific knowledge.

A significant outcome of our study is the inconsistency among evaluation metrics in determining a clear best-performing ASR system. While various metrics have been employed, they exhibited divergent rankings, challenging their ability to comprehensively assess system performance. These discrepancies underscore the necessity to explore alternative evaluation approaches for both intrinsic and downstream evaluations tailored to the task at hand.

\section*{Acknowledgment}

This work was financially supported by the DIETS project financed by the Agence Nationale de la Recherche (ANR) under contract ANR-20-CE23-0005.




\bibliographystyle{IEEEtran}
\bibliography{mybib}

@inproceedings{galliano2006corpus,
  title={{Corpus description of the ESTER Evaluation Campaign for the Rich Transcription of French Broadcast News}},
  author={Galliano, Sylvain and Geoffrois, Edouard and Gravier, Guillaume and Bonastre, Jean-Fran{\c{c}}ois and Mostefa, Djamel and Choukri, Khalid},
  booktitle={International Conference on Language Resources and Evaluation (LREC)},
  year={2006}
}

@article{babu2021xls,
  title={{XLS-R: Self-supervised cross-lingual speech representation learning at scale}},
  author={Babu, Arun and Wang, Changhan and Tjandra, Andros and Lakhotia, Kushal and Xu, Qiantong and Goyal, Naman and Singh, Kritika and von Platen, Patrick and Saraf, Yatharth and Pino, Juan and others},
  year={2021}
}

@inproceedings{sennrich2016neural,
  title={Neural Machine Translation of Rare Words with Subword Units},
  author={Sennrich, Rico and Haddow, Barry and Birch, Alexandra},
  booktitle={54th Annual Meeting of the Association for Computational Linguistics},
  year={2016},
  organization={Association for Computational Linguistics (ACL)}
}

@inproceedings{kudo2018subword,
  title={Subword Regularization: Improving Neural Network Translation Models with Multiple Subword Candidates},
  author={Kudo, Taku},
  booktitle={56th Annual Meeting of the Association for Computational Linguistics},
  year={2018}
}

@inproceedings{schuster2012japanese,
  title={Japanese and korean voice search},
  author={Schuster, Mike and Nakajima, Kaisuke},
  booktitle={2012 IEEE international conference on acoustics, speech and signal processing (ICASSP)},
  year={2012},
  organization={IEEE}
}

@inproceedings{galliano2009ester,
  title={{The ESTER 2 evaluation campaign for the rich transcription of French radio broadcasts}},
  author={Galliano, Sylvain and Gravier, Guillaume and Chaubard, Laura},
  booktitle={Tenth Annual Conference of the International Speech Communication Association},
  year={2009}
}

@inproceedings{esteve2010epac,
  title={{The EPAC corpus: manual and automatic annotations of conversational speech in French broadcast news}},
  author={Esteve, Yannick and Bazillon, Thierry and Antoine, Jean-Yves and B{\'e}chet, Fr{\'e}d{\'e}ric and Farinas, J{\'e}r{\^o}me},
  booktitle={International Conference on Language Resources and Evaluation (LREC)},
  year={2010}
}

@inproceedings{gravier2012etape,
  title={{The ETAPE corpus for the evaluation of speech-based TV content processing in the French language}},
  author={Gravier, Guillaume and Adda, Gilles and Paulsson, Niklas and Carr{\'e}, Matthieu and Giraudel, Aude and Galibert, Olivier},
  booktitle={International Conference on Language Resources and Evaluation (LREC)},
  year={2012}
}

@inproceedings{giraudel2012repere,
  title={{The repere corpus: a multimodal corpus for person recognition}},
  author={Giraudel, Aude and Carr{\'e}, Matthieu and Mapelli, Val{\'e}rie and Kahn, Juliette and Galibert, Olivier and Quintard, Ludovic},
  booktitle={International Conference on Language Resources and Evaluation (LREC)},
  year={2012}
}

@misc{ravanelli2021speechbrain,
  title={{SpeechBrain}: A General-Purpose Speech Toolkit},
  author={Mirco Ravanelli and Titouan Parcollet and Peter Plantinga and Aku Rouhe and Samuele Cornell and Loren Lugosch and Cem Subakan and Nauman Dawalatabad and Abdelwahab Heba and Jianyuan Zhong and Ju-Chieh Chou and Sung-Lin Yeh and others},
  year={2021}
}

@inproceedings{evain2021task,
  title={Task agnostic and task specific self-supervised learning from speech with lebenchmark},
  author={Evain, Sol{\`e}ne and Nguyen, Manh Ha and Le, Hang and Boito, Marcely Zanon and Mdhaffar, Salima and Alisamir, Sina and Tong, Ziyi and Tomashenko, Natalia and Dinarelli, Marco and Parcollet, Titouan and others},
  booktitle={Thirty-fifth Conference on Neural Information Processing Systems (NeurIPS 2021)},
  year={2021}
}

@inproceedings{evain2021benchmark,
  TITLE = {{LeBenchmark: A Reproducible Framework for Assessing Self-Supervised Representation Learning from Speech}},
  AUTHOR = {Evain, Sol{\`e}ne and Nguyen, Ha and Le, Hang and Zanon Boito, Marcely and Mdhaffar, Salima and Alisamir, Sina and Tong, Ziyi and Tomashenko, Natalia and Dinarelli, Marco and Parcollet, Titouan and Allauzen, Alexandre and Est{\`e}ve, Yannick and Lecouteux, Benjamin and Portet, Fran{\c c}ois and Rossato, Solange and Ringeval, Fabien and Schwab, Didier and Besacier, Laurent},
  BOOKTITLE = {{INTERSPEECH 2021: Conference of the International Speech Communication Association}},
  YEAR = {2021},
}

@inproceedings{reimers2019sentence,
  title={{Sentence-BERT: Sentence Embeddings using Siamese BERT-Networks}},
  author={Reimers, Nils and Gurevych, Iryna},
  booktitle={Conference on Empirical Methods in Natural Language Processing and the 9th International Joint Conference on Natural Language Processing (EMNLP-IJCNLP)},
  year={2019}
}

@inproceedings{martin2020camembert,
  title={{CamemBERT: a Tasty French Language Model}},
  author={Martin, Louis and Muller, Benjamin and Su{\'a}rez, Pedro Javier Ortiz and Dupont, Yoann and Romary, Laurent and De La Clergerie, {\'E}ric Villemonte and Seddah, Djam{\'e} and Sagot, Beno{\^\i}t},
  booktitle={58th Annual Meeting of the Association for Computational Linguistics},
  year={2020}
}

@inproceedings{baneras2023hats,
  title={{HATS: An Open data set Integrating Human Perception Applied to the Evaluation of Automatic Speech Recognition Metrics}},
  author={Ba{\~n}eras-Roux, Thibault and Wottawa, Jane and Rouvier, Mickael and Merlin, Teva and Dufour, Richard},
  booktitle={Text, Speech and Dialogue},
  year={2023}
}

@inproceedings{roux2022qualitative,
  title={{Qualitative Evaluation of Language Model Rescoring in Automatic Speech Recognition}},
  author={Ba{\~n}eras-Roux, Thibault and Rouvier, Mickael and Wottawa, Jane and Dufour, Richard},
  booktitle={Interspeech 2022},
  year={2022}
}

@inproceedings{kim2021semantic,
  author={Suyoun Kim and Abhinav Arora and Duc Le and Ching-Feng Yeh and Christian Fuegen and Ozlem Kalinli and Michael L. Seltzer},
  title={{Semantic Distance: A New Metric for ASR Performance Analysis Towards Spoken Language Understanding}},
  year={2021},
  booktitle={Interspeech},  doi={10.21437/Interspeech.2021-1929}
}

@inproceedings{zhang2019bertscore,
  title={BERTScore: Evaluating Text Generation with BERT},
  author={Tianyi Zhang and Varsha Kishore and Felix Wu and Kilian Q. Weinberger and Yoav Artzi},
  booktitle={International Conference on Learning Representations},
  year={2020},
}

@inproceedings{mdhaffar2019qualitative,
  title={Qualitative Evaluation of ASR Adaptation in a Lecture Context: Application to the PASTEL Corpus.},
  author={Mdhaffar, Salima and Est{\`e}ve, Yannick and Hernandez, Nicolas and Laurent, Antoine and Dufour, Richard and Quiniou, Solen},
  booktitle={Interspeech},
  year={2019}
}

@inproceedings{kudo2018sentencepiece,
  title={SentencePiece: A simple and language independent subword tokenizer and detokenizer for Neural Text Processing},
  author={Kudo, Taku and Richardson, John},
  booktitle={Conference on Empirical Methods in Natural Language Processing: System Demonstrations},
  year={2018}
}

@inproceedings{singh2021comparative,
  title={Comparative Study of Different Tokenization Strategies for Streaming End-to-End ASR},
  author={Singh, Sachin and Gupta, Ashutosh and Maghan, Aman and Gowda, Dhananjaya and Singh, Shatrughan and Kim, Chanwoo},
  booktitle={2021 IEEE Automatic Speech Recognition and Understanding Workshop (ASRU)},
  year={2021},
  organization={IEEE}
}

@article{baevski2020wav2vec,
  title={wav2vec 2.0: A framework for self-supervised learning of speech representations},
  author={Baevski, Alexei and Zhou, Yuhao and Mohamed, Abdelrahman and Auli, Michael},
  journal={Advances in neural information processing systems},
  year={2020}
}

@article{hsu2021hubert,
  title={Hubert: Self-supervised speech representation learning by masked prediction of hidden units},
  author={Hsu, Wei-Ning and Bolte, Benjamin and Tsai, Yao-Hung Hubert and Lakhotia, Kushal and Salakhutdinov, Ruslan and Mohamed, Abdelrahman},
  journal={IEEE/ACM Transactions on Audio, Speech, and Language Processing},
  year={2021},
  publisher={IEEE}
}

\vspace{12pt}
\end{document}